\documentclass[runningheads]{llncs}
\usepackage{graphicx}

\usepackage{booktabs}
\usepackage{tabularx}
\usepackage{xcolor}

\usepackage{amsmath}
\usepackage{amssymb}
\usepackage[algoruled]{algorithm2e}

\usepackage[misc,geometry]{ifsym}

\newcommand{\ignore}[1]{}

\begin{document}

\title{A Study of Fitness Gains in Evolving Finite State Machines
}


\author{Gábor Zoltai \Letter\inst{1}\orcidID{ 0000-0002-5762-0628}  \and Yue Xie\inst{2}\orcidID{0000-0002-7959-4563} \and Frank Neumann\inst{1} \orcidID{ 0000-0002-2721-3618}}

\authorrunning{Zoltai et al.}

\institute{Optimisation and Logistics, School of Computer and Mathematical Sciences, University of Adelaide, Australia\\
\email{
gabor.zoltai@adelaide.edu.au \Letter, frank.neumann@adelaide.edu.au}\\
\and Bio-Inspired Robotics Laboratory,
Department of Engineering, \\University of Cambridge, United Kingdom
\email{yx388@cam.ac.uk}}


\maketitle              

\begin{abstract}
Among the wide variety of evolutionary computing models, Finite State Machines (FSMs) have several attractions for fundamental research.  They are easy to understand in concept and can be visualised clearly in simple cases.  They have a ready fitness criterion through their relationship with Regular Languages.  They have also been shown to be tractably evolvable, even up to exhibiting evidence of open-ended evolution in specific scenarios. In addition to theoretical attraction, they also have industrial applications, as a paradigm of both automated and user-initiated control.
Improving the understanding of the factors affecting FSM evolution has relevance to both computer science and practical optimisation of control.
We investigate an evolutionary scenario of FSMs adapting to recognise one of a family of Regular Languages by categorising positive and negative samples, while also being under a counteracting selection pressure that favours fewer states.  The results appear to indicate that longer strings provided as samples reduce the speed of fitness gain, when fitness is measured against a fixed number of sample strings.  We draw the inference that additional information from longer strings is not sufficient to compensate for sparser coverage of the combinatorial space of positive and negative sample strings.

\end{abstract}

\section{Introduction}

In all areas of computing inspired by biological evolution, the relationship between the parameters of selection and increases of fitness measures over time is an important factor. This applies to genetic programming, evolutionary computation, evolutionary artificial intelligence, or artificial life.  In genetic programming, this relationship has been studied with respect to the speed at which a desired solution can be derived \cite{ref_roostapour}.  In artificial life, this relationship is connected to the arising of complexity (being an enabler of fitness) \cite{ref_standish}.  For constrained optimisation problems, the rate of change of selection parameters over time has been seen to enhance the performance of genetic algorithms \cite{ref_kazarlis}.  Diversity, which can be seen as another aspect of selection (as a result of the severity of selection) has been extensively studied for its effects on fitness gains \cite{ref_gabor}.  We investigate an example of the relationship between selection parameters and a fitness measure's increase over generations.

\subsection{Background}

Understanding the relationship between selection function and adaptive performance benefits from simplified models, such as the constructs of automata theory.  For the levels of automata from Finite State Machines (FSMs) up to deterministic Turing Machines, successful evolution to recognise formal languages has been reported \cite{ref_naidoo}.

Finite State Machines (FSMs, also known as deterministic finite automata or DFAs) are one of the simplest such classes of entities.  They have several advantages as a model of study:

\begin{itemize}
  \item They are simple in concept;
  \item their structure can be communicated using clear diagrams (for low numbers of states);
  \item their complexity can be measured unambiguously; and
  \item measures of how well an FSM recognises a regular language provide readily comprehensible fitness metrics.
\end{itemize}

For these reasons, FSMs have been a viable research tool in constructing simplified analogies reflecting aspects of biological evolution.  For example, Rasek et al. \cite{ref_rasek} have proposed a definition of FSM-species and studied their formation.  In another study, Moran and Pollack \cite{ref_moran} have created simple ecosystem models of lineages of FSMs interacting in competitive and cooperative modes. 
 They found that some results exhibit indications of open-endedness, i.e. the absence of an apparent upper bound to evolving complexity.

However, seemingly unbounded increases in complexity of evolving computational entities is the exception, rather than the rule.  The review of the literature by Packard et al. \cite{ref_packard} states the realisation that evolutionary simulations and genetic algorithms tend to approach plateaus of complexity. In other words,
the complexity of evolved solutions to a problem posed tends to level out as generations succeed each other.  Most commonly, a state is reached where few novel solutions of higher complexity are generated by such algorithms.

\subsection{Our contribution}

On the one hand, the literature documents plateaus of evolution under fixed selection parameters.  On the other, there are examples of (at least an impression of) open-endedness with more complex regimes such as co-evolution of three or more interacting lineages.  This suggests a need for better understanding of the influence of the selection regime on the evolution of FSMs, especially in the case of several simultaneous selection pressures.

In order to investigate this, we constructed an experimental setup using two simple selection pressures acting in opposite directions:
\begin{itemize}
    \item maximise the correct recognition of strings in or out of a language, and
    \item minimise the number of states in the FSMs.
\end{itemize}
We use this setup to study the effect of an aspect of selection on fitness gain.

A good metric for the complexity of regular languages is their "state complexity" \cite{ref_yu}.  Unlike other proposed measures (see \cite{ref_ehrenfeucht}), state complexity relates directly to the resources needed for recognition. The family of languages used in this study has one language for each state complexity value. This provides  control over the selection pressure promoting the addition of states over generations.

The remainder of the paper is organised as follows: Section~\ref{sec:pre} provides definitions for the key concepts we have combined in our study, and the way these experiments have applied these ideas.  Section~\ref{sec:alg} presents the algorithms for creation of sample sets, managing the process of evolution, and the mutation operator.  Section~\ref{sec:exp} describes our experiments and their results.  Section~\ref{sec:discuss} includes discussion and interpretation of the results.  Finally, section~\ref{sec:conc} presents our conclusions and proposes future work.

\section{Preliminaries}
\label{sec:pre}

We summarise the elements we have drawn on in  the experimental configuration: FSMs, the \textit{universal witness} family of regular languages, and the way we have combined conflicting selection pressures.

\subsection{Finite State Machines}
A Finite State Machine is a 5-tuple $(Q, \Sigma, \delta, q_0, F)$, where
\begin{itemize}
    \item $Q$ is a set of states;
    \item $\Sigma$ is a set of symbols used as an input alphabet;
    \item $\delta: Q \times \Sigma \rightarrow Q$ is a transition function from one state to another depending on the next input symbol;
    \item $q_0$ is the starting state; and
    \item $F \subseteq Q$ is the subset of accepting states
\end{itemize}

The transition function $\delta$ can be extended to describe what state the machine transitions to if given each symbol of a string in succession.  Using $\epsilon$ for the empty string, and $cs$ to denote the symbol $c$ followed by the trailing sub-string $s$, we can define the extended transition function $\hat{\delta}: Q \times \Sigma^\ast \rightarrow Q$:

\begin{align}
\hat{\delta}(q, \epsilon) &= q \\
\hat{\delta}(q, c s) &= \hat{\delta}(\delta(q,c), s)
\end{align}

An FSM "accepts" a string $s$ if the extended transition function maps it from the starting state to an accepting state, i.e. the following is true:

\begin{equation}
\hat{\delta}(q_0, s) \in F
\end{equation}

The set of strings accepted by an FSM is termed its language; the machine is said to "recognise" the language.  For a machine $\mathcal{M}=(Q, \Sigma, \delta, q_0, F)$, we define its language $L_\mathcal{M}$ as:

\begin{equation}
L_\mathcal{M} = \{ s \in \Sigma^\ast \mid \hat{\delta}(q_0, s) \in F \}
\end{equation}

The languages recognisable by FSMs are the "regular" languages, i.e. those that can be described by a regular expression \cite{ref_sipser}.  Each regular language is recognised by a set of FSMs which are equivalent in that for any string, each of them produces the same result of acceptance or rejection.  The lowest number of states of any FSM recognising a regular language is termed the "state complexity" of the language, as explained by Yu \cite{ref_yu}.  For an algorithm to compute the state complexity from any equivalent FSM, see \cite{ref_hopcroft}.

\subsection{The $U_n$ languages as drivers of complexity}
To be able to run experiments with a range of language complexities, we looked for a family of languages that differed primarily in their complexity only, and preferably only included one member language for each state complexity value.  Both or these criteria are met by the "universal witness" languages described by Brzozowski \cite{ref_brzozowski}, which we use here.  This is a family (or "stream") of languages $U_n$, which has one member for any $n <= 3$, which are examples ("witnesses") of the upper bounds of state complexity in a number of theorems concerning the complexity of FSMs. In particular, it has the property that $U_n$ cannot be recognised by an FSM of less than $n$ states.

\begin{figure}[t]
    \includegraphics[width=\textwidth]{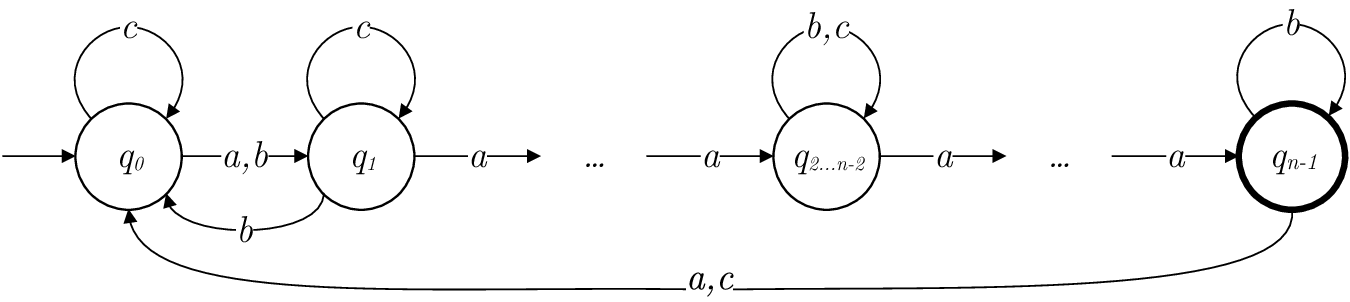}
    \caption{FSM structure for $\mathcal{U}_n$ (based on \cite{ref_brzozowski})}
    \label{fig:u-language-fsm}
\end{figure}

Each of the $U_n$ languages has a corresponding FSM $\mathcal{U}_n$ that accepts it.  The general structure of these FSMs is shown in Figure \ref{fig:u-language-fsm}. The formal definition is as follows:

\subsubsection{Definition of $\mathcal{U}_n$} - for $n \ge 3$, the FSM $\mathcal{U}_n$ is a quintuple $(Q, \Sigma, \delta, q_0, F)$, where $Q = \{q_0, q_1, ..., q_{n-1}\}$ is the set of states, $\Sigma = \{a, b, c\}$ is the alphabet, $q_0$ is the initial state, $F = \{q_{n-1}\}$ is the set containing the one accepting state, and $\delta$ is:\\
    $\delta(q_i, a) = q_{(i+1) \mod n}$; \\
    $\delta(q_0, b) = q_1$; $\delta(q_1, b) = q_0$; $\delta(q_i, b) = q_i$ for $i \notin \{0, 1\}$;\\
    $\delta(q_{n-1}, c) = q_0$; $\delta(q_i, c) = q_i$ for $i \neq n-1$.

The proportion of strings from the ternary alphabet that are members of the $U_n$ languages decreases with increasing $n$, as shown in Table \ref{tab:ulangstats}.

\begin{table}[t]
    \caption{Member strings of some $U_n$ languages as percentage of all ternary strings up to a length}
    \label{tab:ulangstats}
    \centering
    \vspace*{5mm}
    \begin{tabular}{
        r
        *4{| r r}
    }
        \toprule
        \multicolumn{1}{c}{Up to} &
        \multicolumn{2}{c}{All strings} &
        \multicolumn{2}{c}{$U_3$} &
        \multicolumn{2}{c}{$U_4$} &
        \multicolumn{2}{c}{$U_5$} \\
        \cmidrule{2-9}
            length&\#&\%&\#&\%&\#&\%&\#&\% \\
        \hline
            7&3,280&100\%&656&20\%&490&15\%&320&10\% \\
            8&9,841&100\%&1,968&20\%&1,452&15\%&1,122&11\% \\
            9&29,524&100\%&5,904&20\%&4,280&15\%&3,616&12\% \\
            10&88,573&100\%&17,714&20\%&12,688&14\%&11,040&13\% \\
            11&265,720&100\%&53,144&20\%&37,874&14\%&32,640&12\% \\
            12&797,161&100\%&159,432&20\%&113,548&14\%&95,042&11\% \\
            13&2,391,484&100\%&478,296&20\%&340,992&14\%&276,016&12\% \\
            14&7,174,453&100\%&1,434,890&20\%&1,024,128&14\%&806,000&11\% \\
            15&21,523,360&100\%&4,304,672&20\%&3,074,490&14\%&2,375,360&11\% \\
            16&64,570,081&100\%&12,914,016&20\%&9,225,836&14\%&7,065,762&11\% \\
        \bottomrule
    \end{tabular}

\end{table}

\subsection{Conflicting objectives: sample set recognition vs state count}

The two objectives we combine in the selection applied in this study are:
\begin{enumerate}
    \item "linguistic fitness" - maximising the correct classification by FSMs of strings as being in or out of a language; and
    \item minimising the number of states of the FSMs.
\end{enumerate}

The first objective favours more states, as more states can embody more complex information about the target language.  The second objective selects for fewer states.  In this way, the two objectives counteract each other.


To quantify the fitness of an FSM as an acceptor of a language, we use its performance in accepting or rejecting strings in two sample sets, positive (strings in the language) and negative (strings not in the language).

In this study, the "linguistic fitness" $F_{A,R}(\mathcal{M})$ of an FSM $\mathcal{M}$ against a set $A$ of strings to be accepted and a set $R$ of strings to be rejected, is the total of correctly accepted and correctly rejected examples:

\begin{equation}
    \begin{split}
        F_{A,R}(\mathcal{M}) = &\lvert \{ a \in A \mid \mathcal{M}(a) \} \rvert \\
        + &\lvert \{ r \in R \mid \neg \mathcal{M}(r) \} \rvert
    \end{split}
\end{equation}

The maximum value of $F_{A,R}$ for an FSM is the total size of the positive and negative reference sets $A$ and $R$.  The fitness gains we study are specifically the increases of this linguistic fitness over time.


A genetic algorithm selecting for multiple objectives needs to decide which of a set of candidates to favour in the reproductive pool, based on how well they satisfy each of the objectives.  As described by Neumann \cite{ref_neumann}, this can be expressed in a "dominance" relationship combining the two objectives.

In this study, we combine the objectives of maximisation of fitness and minimisation of states.  The former is measured by a function $F_{A,R}$ ($A$ and $R$ being the positive and negative sample sets), the latter by the number of states, $C$.

This paper uses the two relations "weakly dominates" (denoted by the symbol $\succcurlyeq$) and "dominates" (denoted by $\succ$) between FSMs X and Y:
\begin{align*}
    X \succcurlyeq Y &\iff F_{A,R}(X) \geq F_{A,R}(Y) \wedge C(X) \leq C(Y)\\
    X \succ Y &\iff (X \succcurlyeq Y) \wedge (F_{A,R}(X) > F_{A,R}(Y) \vee C(X) < C(Y))
\end{align*}

Informally, X weakly dominates Y if it is no worse than Y on either fitness or complexity, and X dominates Y if it is no worse on either fitness or number of states, but better on at least one of the two.

\section{Algorithms}
\label{sec:alg}

\subsection{Sample set generation methods}
We use two methods of generating sample sets of positive and negative examples. In both methods, the positive and negative sets contain 500 strings each, for a total of 1,000 strings. The two methods labelled $Bss$ and $Rle_n$, are defined below. 

The reason for equal numbers of positive and negative examples (regardless of the percentage of strings of a given length that are in versus out of the language) is technical.  An FSM of a single state can either accept all strings or rejects all strings.  Its fitness score therefore would be the number of positive and negative samples respectively, but its complexity is the same in either case.  Using the same number of positive and negative examples provides a common baseline of performance for minimal complexity.

\subsubsection{$Bss$ method}
The $Bss$ method (for "balanced short strings") generates positive and negative sets that have short strings, with the lengths of positive and negative examples being balanced, i.e. neither the positive nor the negative subset favours longer strings.  The sample sets are generated by Algorithm \ref{alg:Bss}.

\begin{algorithm}[t]
    \label{alg:Bss}
    \caption{Sample set generation: $Bss$ method}
    \SetKwInOut{KwIn}{Input}
    \SetKwInOut{KwOut}{Output}
    \KwIn{$\mathcal{U}$, FSM recognising the target language\\
        $D$, total size of positive and negative sample sets to generate}
    \KwOut{$A$, set of positive example strings\\
        $R$, set of negative example strings}

    $A \leftarrow \{\}$\;
    $R \leftarrow \{\}$\;
    \For{all strings $s$ on the alphabet of $\mathcal{U}$, in alphabetic order}{
        \uIf{$ \mathcal{U}(s) \land \lvert A \rvert \le \lvert R \rvert$}{
            $A \leftarrow A \cup \{s\}$ \;
        }
        \ElseIf{$ \neg(\mathcal{U}(s)) \land \lvert A \rvert \ge \lvert R \rvert$}{
            $R \leftarrow R \cup \{s\}$ \;
        }
        \If{$\lvert A \rvert + \lvert R \rvert \ge D$}{
            break\;
        }
    }
\end{algorithm}

\subsubsection{$Rle_n$ method}
To investigate the effect of longer sample strings, we devised the $Rle_n$ method (for "Random, less than or equal to $n$"). 
This constructs positive and negative sample sets of equal size, from strings drawn uniformly at random from all strings over the alphabet of the target language, of length zero through $n$. The $Rle_n$ method is implemented by by Algorithm \ref{alg:Rle}.

\begin{algorithm}[t]
    \label{alg:Rle}
    \caption{Sample set generation: $Rle_n$ method}
    \SetKwInOut{KwIn}{Input}
    \SetKwInOut{KwOut}{Output}
    \KwIn{$n$, maximum length of strings to include in the sample sets\\
        $\mathcal{U}$, FSM recognising the target language\\
        $D$, total size of positive and negative sample sets to generate}
    \KwOut{$A$, set of positive example strings\\
        $R$, set of negative example strings}

    $A \leftarrow \{\}$\;
    $R \leftarrow \{\}$\;

    $C \leftarrow \lvert$alphabet of $\mathcal{U} \rvert$\;

    $p \leftarrow$ distribution of 0 to $C$, weighted by number of strings of that length\;
    \While{$\lvert A \rvert + \lvert R \rvert < D$}{
        $L \leftarrow $ random choice of length from the weighted distribution $p$\;
        $s \leftarrow$ random string of length $L$ from the alphabet of $\mathcal{U}$\;

        \If{$s \notin A \cup R$}{
            \uIf{$\mathcal{U}(s) \land \lvert A \rvert < D/2 $}{
                $A \leftarrow A \cup \{s\}$ \;
            }
            \ElseIf{$\neg\mathcal{U}(s) \land \lvert R \rvert < D/2 $}{
                $R \leftarrow R \cup \{s\}$ \;
            }
        }
    }
\end{algorithm}

\subsection{The evolution algorithm}
The study of complexity in multi-objective genetic programming by Neumann \cite{ref_neumann} describes the SMO-GP algorithm (Simple Multi-Objective Genetic Programming).  It starts with a population of one individual.  In each iteration, SMO-GP selects one member of the population, applies a mutation operation, and then either replaces all dominated individuals with the (single) new mutant, or drops the mutant if it does not dominate any existing members.

For this study, we adapted this algorithm is adapted to evolve FSMs starting with the simplest possible FSM, which we call $\mathcal{N}$, see Algorithm \ref{alg:smogp}.
$\mathcal{N}$ is an FSM of a single non-accepting state, with the transition function mapping all inputs back to that state.  This is in a sense the simplest possible FSM, and corresponds to the empty language, as it rejects all strings.  (An FSM accepting all strings, i.e. accepting $\ast{\Sigma}$, would be equally simple.)  The formal definition of such an FSM for a ternary alphabet is:

\begin{equation}
    \mathcal{N} = (\{q_0\},\{a,b,c\},\{((q_0,a), q_0), ((q_0,b), q_0), ((q_0,c), q_0)\}, q_0, \emptyset)
\end{equation}

Note that half the 1,000 sample strings will not be elements of the language, so the FSM $\mathcal{N}$ by rejecting them will achieve a linguistic fitness score of 500.  The only way for a new mutant to enter the population is to weakly dominate an existing FSM in the population.  This can come about by having higher fitness or fewer states, but no FSM can have less states than $\mathcal{N}$.  Therefore, all other FSMs that will become part of the population will score at least 500.

In some preliminary work, we also experimented with starting evolution towards $U_n$ not with $\mathcal{N}$, but instead with the result of evoluition over a set number of generations toward $U_{n-1}$.  This approach did not produce any notable patterns correlating the starting point with improved speed of adaptation to higher fitness, and we did not pursue it further.

\begin{algorithm}[t]
\label{alg:smogp}
    \caption{Multi-objective Evolution Algorithm for FSMs}
    \SetKwInOut{KwIn}{Input}
    \SetKwInOut{KwOut}{Output}
    \KwIn{$G$, then number of iterations}
    \KwOut{$P$, the final set of FSMs in the population after $G$ generations}

    \BlankLine
 
    $q_0 \leftarrow $an initial FSM state\;
    $\mathcal{N} \leftarrow (\{q_0\},\{a,b,c\},\{((q_0,a), q_0), ((q_0,b), q_0), ((q_0,c), q_0)\}, q_0, \emptyset)$\;

    \BlankLine
    $P \leftarrow \{\mathcal{N}\}$\;
    \BlankLine

    \For{$G$ iterations}{
        $M \leftarrow $ uniformly random choice from $P$\;
        
        \tcp{Pois denotes the Poisson distribution with parameter $\lambda = 1$.}
        \For{$1 + $Pois$(1)$ iterations} { 
            $M \leftarrow M$ modified by mutation operation\;
        }

        \If{$\nexists i \in P \vert i \succ M$} {
            $P \leftarrow ( P \cup \{ M \} ) \setminus \{ j \in P \vert M \succcurlyeq j \}$\;
        }
    }

\end{algorithm}

\subsection{The mutation operation}
The mutation operation used on FSMs selects one random state for the source of an arc, and a random input symbol from the alphabet for the label of the arc.  A target state is then selected; half the time, this is an existing state of the FSM, otherwise it is a new state which has all outgoing arcs leading back to itself.

The FSM is then modified so that the arc from the source state, labelled with the chosen input symbol, is set to lead to the chosen target state.

Lastly, a random state is chosen and changed from accepting to non-accepting and vice versa.  For details, see Algorithm \ref{alg:mutation}.

\begin{algorithm}[t]
    \label{alg:mutation}
    \caption{Mutation operation}
    \SetKwInOut{KwIn}{Input}
    \SetKwInOut{KwOut}{Output}
    \KwIn{$M = (Q, \Sigma, \delta, q_0, F)$, the FSM to be mutated}
    \KwOut{$M$, the input FSM modified}

    $q \leftarrow $ uniformly random choice from $Q$\;
    $c \leftarrow $ uniformly random choice from $\Sigma$\;
    \uIf {uniformly random choice of \{existing, new\} = existing} {
        $t \leftarrow $ uniformly random choice from $Q$\;
    }
    \Else {
        $t \leftarrow $ a new FSM state $\notin Q$\;
        $Q \leftarrow Q \cup \{t\}$\;
    }
    $\delta(q,c) \leftarrow t$\;
    \BlankLine
    $r \leftarrow $ uniformly random choice from $Q$\;

    \uIf{uniformly random choice of \{accepting, rejecting\} = rejecting} {
        $F \leftarrow F \setminus \{r\}$\;
    }
    \Else {
        $F \leftarrow F \cup \{r\}$\;
    }

\end{algorithm}

\section{Experimental results}
\label{sec:exp}

\begin{figure}[t]
    \centering
    \vspace{-0.5cm}
     \includegraphics[scale=0.23]{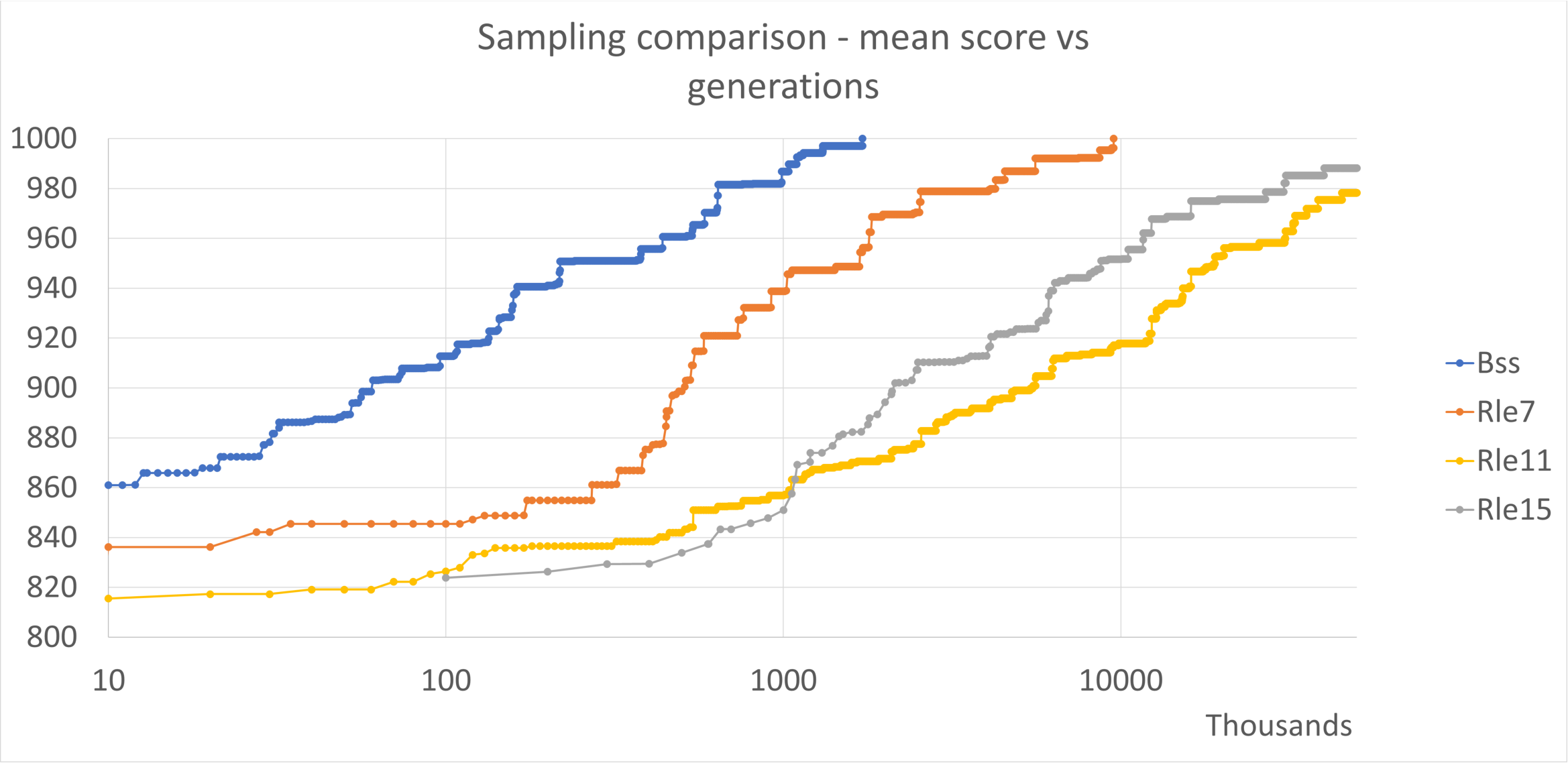}

    \caption{Evolving to $U_3$: Comparison of sampling styles}
    \label{fig:U3_comparison_chart}
    \vspace{-0.5cm}
\end{figure}

Table \ref{tab:scorecomp} shows the mean fitness scores observed after given numbers of generations, under four different methods of generating sample sets for the $U_3$ language, across 30 runs with different randomisation seeds.  Figure \ref{fig:U3_comparison_chart} shows a more detailed set of observations for the four methods.

\begin{figure}[t]
    \centering
    \vspace{-0.5cm}
    \includegraphics[width=\textwidth]{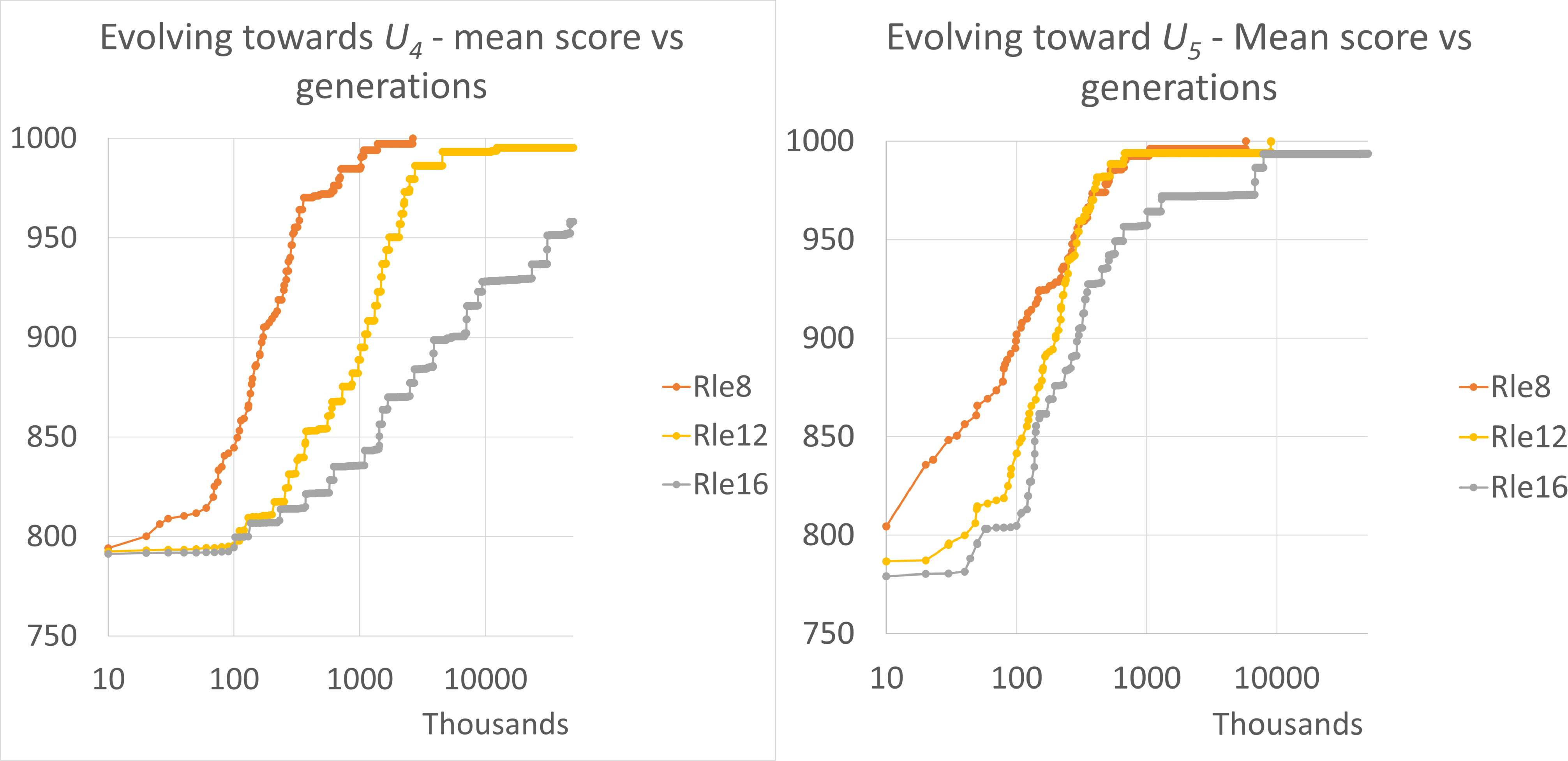}
    \caption{Evolving to $U_4$ and $U_5$: Comparison of sampling styles}
    \vspace{-0.5cm}
    \label{fig:U4_U5_comparison_chart}
\end{figure}

The relative position of the data series in Figure \ref{fig:U3_comparison_chart} made us curious about whether a general trend might exist for using longer strings in sample sets to lead to slower evolution of better recognisers.
We repeated the experiment for $U_4$ and $U_5$; results are shown in Figure \ref{fig:U4_U5_comparison_chart}.
Note that the $Rle$ sampling was adjusted for $U_4$ and $U_5$.  In the $U_4$ experiment, we generated 500 positive and 500 negative sample strings, for each of $Rle_7$, $Rle_{11}$, and $Rle_{15}$.  However, the languages $U_4$ and $U_5$ do not contain 500 strings of at most 7 characters.  In order to create comparable sample sets for $U_4$ and $U_5$, we used $Rle_8$, $Rle_{12}$, and $Rle_{16}$.

\begin{table}
    \caption{Score comparison for evolving $U_3$ from $\mathcal{N}$  (mean of 30 runs with different randomisation seeds)}
    \label{tab:scorecomp}
    \centering
    \vspace*{5mm}
    \begin{tabularx}{\textwidth}{r *{4}{>{\raggedleft\arraybackslash}X}}
        \toprule
        \multicolumn{1}{c}{} &
        \multicolumn{4}{c}{Mean score} \\
        \cmidrule{2-5}
            Generation&$Bss$&$Rle_7$&$Rle_{11}$&$Rle_{15}$ \\
        \hline
            100,000 &912.67&845.47&826.37&823.83 \\
            200,000 &941.03&854.90&836.57&826.33 \\
            500,000 &960.63&898.60&841.90&833.87 \\
            1,000,000 &986.80&938.73&856.80&851.03 \\
            2,000,000 &1000.00&969.60&871.70&894.30 \\
            5,000,000 &1000.00&986.97&898.97&923.57 \\
            10,000,000 &1000.00&1000.00&917.87&951.63 \\
        \bottomrule
        \\
    \end{tabularx}
\end{table}

\section{Discussion}
\label{sec:discuss}

A pattern of higher $n$ used in $Rle_n$ sampling leading to slower improvement of recognition fitness is seen in all three Figures summarising the generated data.  This observation suggests that selection by using longer strings provides less relevant information towards recognising the language than selection by shorter strings.  This is perhaps counter-intuitive. 
 Longer sample strings of necessity embed more information about the language in question, on a per string basis.

A possible explanation may be that as we increase $n$ in $Rle_n$, a fixed-size sample set (e.g. 500 positive and 500 negative strings) is a smaller percentage of all strings of length up to $n$.  For a long string, a larger set of its leading substrings is likely to be missing from the sample set. So, a mutant FSM may randomly be correct in its categorisation of the long string, while not correctly categorising its prefixes.  This implies that a gain of fitness under these circumstances is less likely to lead to further gains of fitness, compared to sample sets that are a more comprehensive coverage of shorter strings in the language.

In other words, the information about the language that is embedded in the sample set is less representative at higher values of $n$ in $Rle_n$.  With less comprehensive coverage of examples for higher $n$ in $Rle_n$, the positive and negative sets of examples may become more akin to random strings rather than providing selection conducive to higher linguistic fitness in the next generation.

Although these experiments only investigated evolution against sample sets of $U_3$ through $U_5$, for further members of the series, growth in the number of FSM states may be an important factor.  As it is known that a perfect recogniser for $U_n$ needs to have at least $n$ states, we would expect that the growth in the number of states would be an important factor in enabling the evolution of better recognition for higher values of $n$ in $U_n$.

Whether this finding transfers to other languages, and if so, whether it may be able to be generalised to other automata or evolutionary scenarios, is not answered by our study.  Even within the $U_n$ languages, we have only investigated low values of $n$. 
 Further, languages with significantly different structures may behave differently. Algorithm \ref{alg:smogp} uses very small populations, so the effect of diversity on our observation is also an open question.
 
A potential application of our observation, if generalised, may be to find better solutions in fewer generations in evolutionary optimisation.

\section{Conclusion}
\label{sec:conc}

Under the experimental scenario we used, extending the sample sets of training strings to longer string lengths tends to slow down the achievement of higher fitness scores.  A fixed size of sample set, drawn from pools of strings that increase in size as maximum length is extended provides less directed selection pressure.  It appears that this is not counteracted by the additional information embedded in longer sample strings, possibly because gains of fitness may occur with a lower likelihood of leading to further gains of fitness, using the same fitness function.

This work was based on observation of the $U_n$ languages, for low values of the state complexity $n$.  It would be interesting to investigate whether these observations hold for:
- $U_n$ with higher values of $n$, or
- for other languages arranged in a sequence of increasing state complexity, or
- when the linguistic fitness function shifts from $U_n$ to $U_{n+1}$.

Another fruitful avenue of further work may be to relax the criteria governing retention of FSMs in the population, and observe the effect of populations with higher diversity interacting with rising string lengths in the sample sets.

\section{Acknowledgements}
This work was supported by the Commonwealth of Australia.
Frank Neumann was supported by the Australian Research Council through grant FT200100536.

%
%
%
\bibliographystyle{MathPhySci}
%

\end{document}